\documentclass[letterpaper, 10 pt, conference]{ieeeconf}  % Comment this line out if you need a4paper

\IEEEoverridecommandlockouts                              % This command is only needed if 
\usepackage[utf8]{inputenc}
\usepackage[T1]{fontenc}
\usepackage{graphicx}
\usepackage{amsmath}
\usepackage{listings}
\usepackage{multirow}
\usepackage[dvipsnames,svgnames,x11names]{xcolor}
\usepackage[final]{changes}
\usepackage{tabularray}
\usepackage[normalem]{ulem}
\usepackage{siunitx}
\usepackage{gensymb}
\usepackage{tablefootnote}
\usepackage{url}
\usepackage{hyperref}
\hypersetup{
    urlcolor=cyan,
    pdftitle={Overleaf Example},
    pdfpagemode=FullScreen,
    }

\title{\LARGE \bf
A Hybrid Approach to Indoor Social Navigation: Integrating Reactive Local Planning and Proactive Global Planning
}

\author{{Arnab Debnath$^1$}, {Gregory J. Stein$^1$} and {Jana Ko{\v{s}}eck{\'a}$^1$}
\thanks{$^{1}$Arnab Debnath, Gregory J. Stein and Jana Ko{\v{s}}eck{\'a} are with the Department of Computer Science, 
George Mason University, 
4400 University Dr, Fairfax, VA, USA {\tt\small \{adebnath, gjstein, kosecka\}@gmu.edu}}}

\begin{document}

\maketitle
\thispagestyle{empty}
\pagestyle{empty}
 \begin{abstract} 
We consider the problem of indoor building-scale social navigation, where
the robot must reach a point goal as quickly as possible without
colliding with humans who are freely moving around. Factors such as varying crowd densities, unpredictable human behavior, and the constraints of indoor spaces add significant complexity to the navigation task, necessitating a more advanced approach. We propose a modular navigation framework that leverages the strengths of both classical methods and deep reinforcement learning (DRL). Our approach employs a global planner to generate waypoints, assigning soft costs around anticipated pedestrian locations, encouraging caution around potential future positions of humans. Simultaneously, the local planner, powered by DRL, follows these waypoints while avoiding collisions. The combination of these planners enables the agent to perform complex maneuvers and effectively navigate crowded and constrained environments while improving reliability. Many existing studies on social navigation are conducted in simplistic or open environments, limiting the ability of trained models to perform well in complex, real-world settings. To advance research in this area, we introduce a new 2D benchmark designed to facilitate development and testing of social navigation strategies in indoor environments.\footnote[2]{Simulator and code: \url{https://github.com/arnabGMU/hybrid_social_nav}} We benchmark our method against traditional and RL-based navigation strategies, demonstrating that our approach outperforms both. % in terms of z, showcasing its superior performance in social navigation tasks.
 \end{abstract}

\section{Introduction}
We consider a robot deployed in an indoor environment where pedestrians are moving around, and given a limited time, the robot needs to reach a point goal without colliding with the pedestrians. 
Mobile robots often need to navigate in environments populated by people, and the robots are expected to
socially react to them while successfully reaching their target. The ability to effectively and safely navigate is critical in applications ranging from service robots in domestic assistance \cite{eirale2022human} and healthcare \cite{holland2021service} settings to delivery robots \cite{hossain2023autonomous} transporting packages or takeout orders in both indoor and outdoor environments. These robots must balance efficiency with social compliance to accomplish their tasks without disrupting the human environment.

Effective indoor social navigation faces multiple significant challenges. Indoor environments may contain clutter, complicating the robot's maneuvering ability. 
Many previous works have focused on open spaces \cite{chen2017decentralized, chen2017socially} or small geometrically simple environments \cite{long2018towards}, often with few or no obstacles, limiting their ability to perform well in more geometrically complex, dynamic environments with a high number of pedestrians, characteristic of household environments. 

%Furthermore, predicting future human behavior, which can significantly impact navigation, is difficult. In the pursuit of developing socially adept mobile robots, advancements have been made in the research concerning human-robot co-navigation. Progress has been observed in both classical planning and learning methodologies for robots capable of managing local interactions with pedestrians moving within crowds.

% Many previous works on social navigation have focused
% on open spaces or simplified environments, often with few or
% no static obstacles. These studies typically focus on a limited
% set of scenarios, such as head-on collisions or corridor
% navigation with a small number of pedestrians, restricting
% their applicability to more complex, dynamic settings. In this
% work, we address the problem of indoor social navigation that presents significant challenges due to the various factors that
% come into play. Indoor environments may contain clutter,
% complicating the robot’s maneuvering ability. Furthermore,
% predicting future human behavior, which can significantly
% impact navigation, is difficult. In the pursuit of develop-
% ing socially adept mobile robots, advancements have been
% made in the research concerning human-robot co-navigation.
% Progress has been observed in both classical planning and
% learning methodologies for robots capable of managing local
% interactions with pedestrians moving within crowds

Traditional methods for dynamic collision avoidance offer valuable attributes such as safety, reliability, and interpretability. However, they often fail to work very effectively in more complex indoor environments. Such approaches, including potential field methods \cite{park2001obstacle} or velocity obstacle methods \cite{van2008reciprocal, van2011reciprocal} typically depend on predefined reactive safety rules. Additionally, methods such as Dynamic Window Approach (DWA) \cite{fox1997dynamic} or Time Elastic Band (TEB) \cite{rosmann2015timed} often involve the repeated solution of optimization control problems. Objective functions of these planners can incorporate various costs or constraints to enhance social compliance, but this incorporation of the cost often involves a complex and time-consuming tuning process due to the variety of human dynamics and social rules.
These approaches face particular challenges when faced with swiftly moving pedestrians and in highly cluttered indoor environments, potentially leading to less efficient maneuvers.

 \begin{figure}[t!]
\centering
\includegraphics[width=0.48\textwidth]{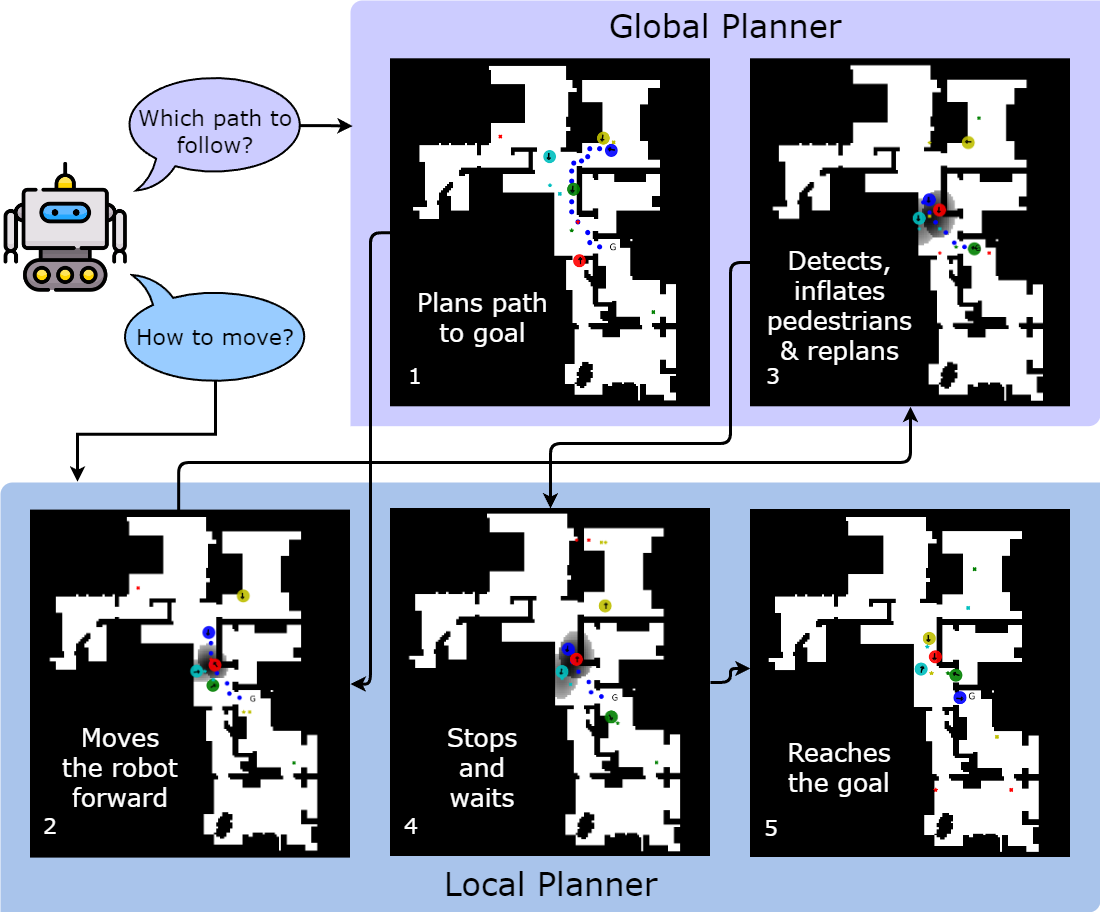}
\vspace{-1.5em}
\caption{Our hybrid planner consists of a non-learned global planner and an RL-based local planner. The global planner generates waypoints to the goal (1) and updates the costmap for replanning when pedestrians are detected (3). The local planner generates control commands that guide the robot (blue) along the waypoints (2, 5). It also exhibits intelligent behavior, such as waiting when the path is blocked by pedestrians (4).}
\vspace{-1em}
\label{fig:title}
\end{figure} 

%Classical approaches require significant computational resources and face challenges in effectively addressing swiftly moving obstacles. 
%Moreover, they often fail to work well given the intricacies of human behaviors, potentially leading to less efficient maneuvers. Despite these limitations,  Classical approaches offer valuable real-world attributes such as safety, reliability, and interoperability. However, a more adaptive approach is needed to deal with diverse and unpredictable social scenarios. 

As a potential solution to these challenges, machine learning (ML) approaches have emerged, such as imitation learning, inverse reinforcement learning, and deep reinforcement learning (DRL) that can leverage data to model complex human behaviors. DRL has proven to be a potentially effective tool that enables robots to learn navigation policies through trial-and-error interactions with their environment, to capture intricate behavioral patterns to achieve effective social navigation \cite{cancelli2022exploiting, perez2021robot, yokoyama2022benchmarking, chen2017socially}. However, DRL suffers from significant sample inefficiency, making DRL approaches frequently myopic. DRL's reliability is often constrained to local planning tasks, which can result in poor performance in more complex or long-term navigation scenarios. Moreover, end-to-end learning approaches (directly mapping perception to motion commands) often generalize poorly in unseen or more challenging environments: i.e., apartments they have never seen before or larger than those seen during training.

%Recent deep learning methods integrating crowd forecasting incur high computational costs and encounter the freezing robot problem due to predicted pedestrian trajectories filling the entire space. DRL's primary drawback lies in its low learning efficiency, especially in navigating large environments with sparse rewards, posing significant challenges for robot training.

To capture the advantages of both classical and learning-based approaches, several hybrid systems have emerged over the years that combine complementary elements from each \cite{xu2021machine, martini2024adaptive, patel2020dynamically}. Indoor social navigation benefits from a hybrid approach, as neither classical nor DRL approaches alone are well-suited to handle the complexities involved which makes the combination of both approaches essential to achieve robust, efficient, and socially-aware navigation. 
%DRL excels at learning collision avoidance in dynamic, cluttered environments with humans exhibiting diverse behaviors, where traditional methods often fall short. However, indoor navigation also demands long-horizon planning, a strength of classical approaches, especially in environments with many dead ends or complex layouts. 
%End-to-end DRL systems tend to struggle in such scenarios, making a combination of both approaches essential to achieve robust, efficient, and socially-aware navigation.

In this work, we introduce a hybrid navigation framework that integrates a classical, non-learned high-level planner with a RL-based local planner for effective social navigation in highly cluttered indoor environments. 
Our high-level planner provides structured guidance by generating waypoints on a 2D map, assigning Gaussian soft costs on the map around pedestrians. Meanwhile, the reactive RL planner manages local navigation by using these waypoints for guidance and dynamically avoiding collisions. This combination leverages the strengths of both methods, improving the performance of navigation through complex indoor environments while avoiding collisions with pedestrians. Fig. \ref{fig:title} illustrates the functionality and some learned behavior of our planner.  
%Our high-level planner offers structured guidance and strategic decision-making, while the RL-based local planner adapts to dynamic, local conditions and handles detailed navigation. This combination leverages the strengths of both methods, significantly enhancing the robustness of our system and enabling reliable navigation through complex indoor environments while avoiding collisions with pedestrians.

%In our hybrid navigation system, the global planner generates waypoints on a 2D map, while the local planner follows these waypoints to ensure safe navigation. Our global planner updates the 2D costmap by anticipating future human positions, without relying on costly tracking or human motion prediction. Meanwhile, the DRL-based local planner, trained in simulation, follows the global waypoints while dynamically avoiding collisions with humans.

Our contributions include the following:

\begin{itemize}
    \item We design a hybrid navigation system for social navigation to boost performance for navigation in challenging indoor environment.
    \item We introduce a global path planner that proactively updates waypoints to prevent collisions with pedestrians.
    \item We propose a reactive DRL-based local planner that, while following the guidance of waypoints from the global planner, avoids imminent collisions by taking safety measures such as stopping or deviating from its original path.
    % It creates a cost map where human positions are inflated with a gaussian noise, according to the pedestrian orientation and distance between the robot and the pedestrian.
    
    \item We introduce a new 2D benchmark for social navigation using ORCA simulator \cite{alonso2013optimal} for pedestrian simulation on 2D maps from iGibson \cite{xia2020interactive}.
     
\end{itemize}
Through extensive experiments, we show that our unified system outperforms both traditional and DRL-based baselines. We further show how each module, local and global planners, contributes to the overall performance through an ablation study. To emulate the challenging conditions of movement through a cluttered indoor environment, we include simulation results in which humans are \emph{uncooperative}---that they behave as if the robot is not there---a realistic environment condition that imposes further challenge on achieving social compliance, particularly in indoor environments where space is constrained and reaction times are limited. By comparing performance across cooperative and uncooperative pedestrian behaviors, we are able to measure the reliability of our navigation system under varying levels of pedestrian interaction difficulty. Our results demonstrate that the system performs consistently better with uncooperative pedestrians, highlighting its capability to handle unpredictable human behavior and further validating its real-world applicability.
\section{Related Work}
\textbf{Non-learned approaches for social navigation}
 In earlier works on collision avoidance \cite{borenstein1989real, fox1997dynamic, thrun2000probabilistic, burgard1999museum, borenstein1990real, borenstein1991vector}, pedestrians were typically treated as static, non-interactive obstacles, overlooking their interactions with other agents in the environment. These approaches often relied on predefined rules to ensure safety \cite{park2001obstacle, van2008reciprocal, van2011reciprocal}, but they demanded substantial computational resources. While such methods are widely adopted in real-world applications due to their reliability and safety, they frequently fall short in capturing the complexities of human behavior, especially in densely populated environments. Cheng et al. \cite{cheng2022dynamic} introduced a dynamic costmap approach to enhance the performance of the DWA planner by assigning costs around dynamic obstacles using a 2D Gaussian function. Building on this concept, our global planner similarly assigns costs around pedestrians.

\textbf{Imitation and Inverse Reinforcement Learning}
More recent approaches have employed various learning-based methods, such as inverse reinforcement learning (IRL), imitation learning, and reinforcement learning, to address the challenges of modeling complex human behaviors. IRL has been utilized to derive reward functions from human trajectory data \cite{kretzschmar2016socially, kim2016socially}, while imitation learning approaches \cite{tai2018socially, long2017deep, liu2018map} use human trajectories to train socially aware navigation policies through supervised learning. However, both rely on large trajectory datasets.
%However, both IRL and imitation learning rely on large, high-quality trajectory datasets, and their lack of adaptability during training hinders reactivity, making them less effective at generalizing to new environments.
\begin{figure*}[t]
    \centering
    \includegraphics[width=\linewidth]{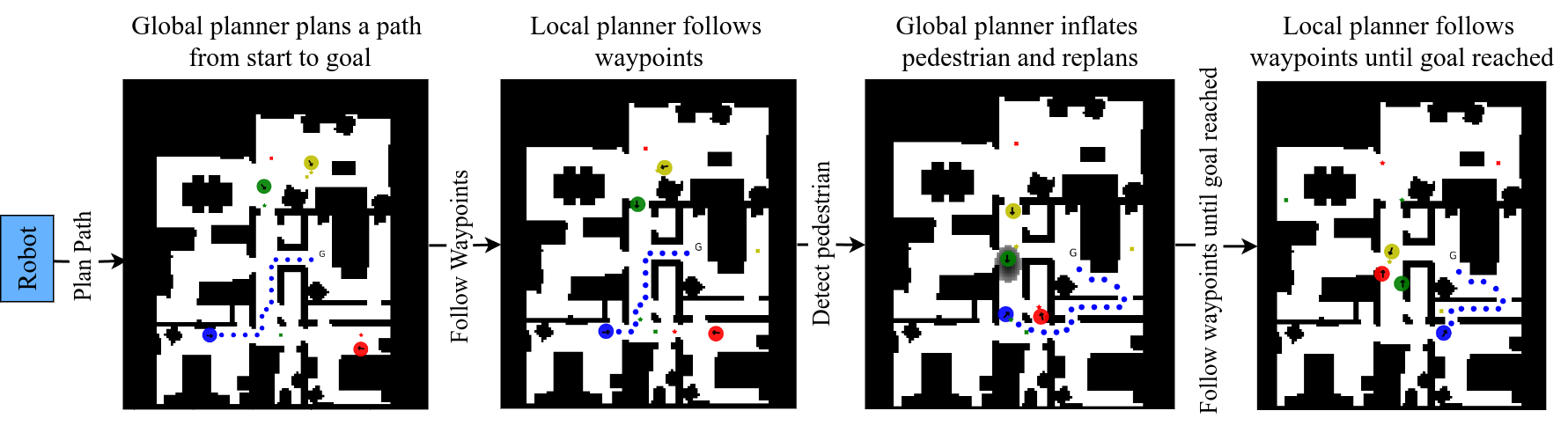}
    \caption{Workflow of our hybrid planner. Initially, the robot generates waypoints at the start of the episode (left). The local planner then follows these waypoints (second from left). As the robot navigates, the global planner detects pedestrians and dynamically replans the path (third from left). Finally, the local planner successfully reaches the goal while avoiding collisions (rightmost). }
    \label{fig:overview}
\vspace{-1em}
    
\end{figure*}

\textbf{Deep Reinforcement Learning}
DRL methods overcome the challenges posed by IRL and imitation learning by directly interacting with the environment using a pre-specified reward function \cite{chen2017decentralized, chen2017socially, long2018towards, everettmotion}. One of the earliest works, CADRL \cite{chen2017decentralized}, and SA-CADRL \cite{chen2017socially} developed a reinforcement learning policy for pairwise collision avoidance between two agents, yet it struggles in crowded environments because they require modeling each unique pairwise interaction individually. To address this, LSTM-RL \cite{everettmotion} was proposed, which considers all pedestrians simultaneously and sequences them based on proximity. SARL \cite{chen2019crowd} further improved performance by combining imitation learning in the early phase of training with an attention-based DRL framework. Despite these advances, end-to-end navigation systems still face challenges in robustness and are less effective for indoor social navigation.
%Despite these advances, end-to-end navigation systems still face challenges in robustness and tend to generalize poorly to novel environments.

\textbf{Hybrid Approaches}
Various hybrid approaches have been proposed to overcome the limitations of the classical and end-to-end learning approaches. Martini et al. \cite{martini2024adaptive} uses DRL to dynamically adjust the cost weights of a Social Window Planner. Raj et al. \cite{raj2024rethinking} proposed a hybrid planner that switches between a learning-based and a geometric planner. Whereas Patel et al. \cite{borenstein1989real} uses DRL to evaluate predicted trajectories of DWA. Kobayashi et al. \cite{kobayashi2023local} used Q-table to dynamically change the cost terms of DWA. Our approach is most similar to the waypoint following planner proposed in \cite{perez2021robot} where the authors combined a sampling based motion planner which generates a fixed a set of waypoints at the beginning of the episode, with a DRL-based local planner. 
%Our local planner differs in terms of observation space and reward shaping which is dicussed in Sec. \ref{section: rl_planner}. Our global planner (Sec. \ref{section: global planner}) updates the costmap according to relative pedestrian position and orientation, and uses replanning when necessary.

\section{Problem Formulation: Social Navigation in Indoor Environment}
\subsection{Overview}

% \begin{figure*}[t]
%     \centering
%     \includegraphics[width=\linewidth]{Figs/overall4.png}
%     %\includegraphics[scale=0.1]{Figs/overall.png}
%     \caption{Workflow of our hybrid planner. Initially, the robot generates waypoints at the start of the episode (left). The local planner then follows these waypoints (second from left). As the robot navigates, the global planner detects pedestrians and dynamically replans the path (third from left). Finally, the local planner successfully reaches the goal while avoiding collisions (rightmost). }
%     \label{fig:overview}
% \vspace{-1em}
    
% \end{figure*}

The agent is deployed at a random location, $(x_i, y_i)$ in an indoor environment where a number of pedestrians are freely moving around. The agent knows the fixed layout of the environment, static obstacles from a 2D map, but does not know the location of the pedestrians until they come into view. We assume that the agent can precisely localize itself within the environment and utilize its limited visibility to create egocentric 2D local maps that show free space and obstacles. If any pedestrian is within the agent's field of view, it can also determine the pedestrian's relative distance and heading. The agent's objective is to reach the goal location $(x_g, y_g)$. The action space for the robot is $(v, \omega)$, where $v$ is the linear velocity and $\omega$ is the angular velocity.

An episode involves the random initialization of source and destination positions for all pedestrians and the robot. A collision with a pedestrian occurs when the robot gets closer than $0.3\si{\meter}$ to any pedestrian, resulting in a failed episode. Wall collisions are detected but do not cause termination of the episode; instead, five consecutive wall collisions result in a random action for all agents. A successful episode requires the agent to get as close as $0.3\si{\meter}$ to the target location. A failed episode occurs if the agent cannot reach the goal within 500 timesteps or collides with pedestrians.
\section{Proposed Approach}

We propose a modular approach to social navigation consisting of two modules: a) the global planner named as Proactive Path Planner (PPP) and b) the reactive RL-based local planner, both described in the next sections.

The global planner (Sec.~\ref{section: global planner}) computes waypoints from the robot position to the target location for the local planner to follow. It anticipates the human's future position and adds a Gaussian soft cost around a location based on their heading, resulting in paths that better avoid future collision. The local planner (RRL) is a reactive RL based planner which can follow the waypoints produced from the global planner while avoiding local collision with humans. 

At the beginning of the episode, the agent uses the Proactive Path Planner (PPP) to generate a set of waypoints from the starting position to the target location. The local planner is then employed to follow these waypoints. As the robot follows the waypoints, it continually receives new observations of the environment. If a human comes in view within 50cm of the robot's waypoints, the robot interacts with the planner again to generate a new set of waypoints. If a collision-free path cannot be achieved from the current state of the environment due to humans making most regions occluded, or if a new potentially collision-free path cannot be traversed within the remaining time, the robot continues on its original path, relying solely on the local planner for collision avoidance. Fig.~\ref{fig:overview} shows an overview of our system.

%----------------------------------------------------------
\subsection{Reactive RL Planner for Local Collision Avoidance}
\label{section: rl_planner}
We train the RL planner using the Soft Actor Critic (SAC) algorithm \cite{haarnoja2018soft} to enable the agent to learn local collision avoidance in the presence of humans. SAC introduces an entropy term into the reward function, encouraging exploration by maximizing entropy. Entropy, in this context, measures the randomness of the policy. A higher entropy means the policy is more stochastic, leading to more exploration. The goal is to find a balance between exploration (trying new things) and exploitation (using known information to maximize rewards). The policy network receives observations and outputs control commands $(v, \omega)$. 

\begin{figure}[!t]
    \centering
    \vspace*{0.5em}
    \includegraphics[width=\linewidth]{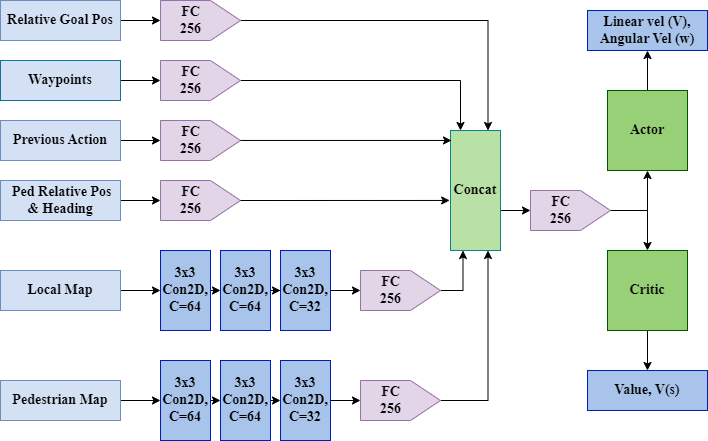}
    \caption{SAC architecture for the reactive RL planner}
    \label{fig:rl_planner}
    \vspace{-1em}
\end{figure}

The observation space for the SAC includes \textit{relative goal position, local map, waypoints, previous action, pedestrian map} and \textit{pedestrian position observations}. The relative goal position represents the goal location's position relative to the agent in polar coordinates. The local map is a $100 \times 100$ egocentric occupancy map, and the pedestrian map marks the cells occupied by pedestrians. The previous action is the action taken by the agent in the immediate last time step. The waypoints are the next five waypoints relative to the agent in polar coordinates, generated by the global planner. Finally, the pedestrian position observations include the relative positions of pedestrians in polar coordinates, along with their relative headings with respect to the agent. All observations are normalized to the range $[-1,1]$. An overview of our architecture is shown in Fig.~\ref{fig:rl_planner}.

Our reward function is designed to incentivize the robot for following the waypoints and penalize it for getting too close to pedestrians or making collisions:
\begin{equation}
    \begin{array}{rcl}
    R & = & R_\text{goal} + R_\text{pedCol} + R_\text{wallCol} + R_\text{wp} + R_\text{timestep} \\  
    & & + R_\text{wpDist} + R_\text{pedAvoid} + R_\text{wpOrient}
    \end{array}
\end{equation}
Here, $R_\text{goal} = 20$ is a sparse reward given for reaching the goal. A penalty of $R_\text{pedCol} = -20$ is applied for colliding with a pedestrian, and $R_\text{wallCol} = -10$ applied for colliding with a wall. The agent receives $R_\text{wp} = 0.8$  if it gets within 0.1 meters of any waypoint. The waypoint distance reward ($R_\text{wpDist}$), waypoint orienation reward ($R_\text{wpOrient}$), pedestrian avoidance reward ($R_\text{pedAvoid}$) and timestep reward ($R_\text{timestep}$) are the dense rewards:
\begin{equation}
    R_\text{wpDist} = w_{1} \times (d_{t-1} - d_{t})
\end{equation}
where $w_1 = 0.3$ and $d_{t-1}$ and $d_t$ are the distance to the next waypoint in timesteps $t-1$ and $t$ respectively;
\begin{equation}
    R_\text{wpOrient} = w_{2} \times (\alpha_{t-1} - \alpha_{t})
\end{equation}
where, $w_2 = 0.3 $ and $\alpha_{t-1}$, $\alpha_t$ are the relative angles to the next waypoint in timestep $t-1$ and $t$ respectively;
\begin{equation}
    R_\text{pedAvoid} = \left\{ 
    \begin{array}{cl}
            -\frac{(d_\text{thresh} - d_\text{ped})} {d_\text{thresh} - d_\text{col}} & : d_\text{ped} \leq d_\text{thresh} \\
            0 & : \mbox{otherwise}
        \end{array} \right.      
\end{equation}
The pedestrian collision avoidance reward is applied when the closest pedestrian is within 1 meter:  i.e $d_\text{ped} < d_\text{thresh} = 1$. $d_\text{col} = 0.3$ refers to the distance at which a collision occurs. 

\noindent ${R_\text{timestep}} = -0.001$ is a small negative reward applied at each timestep to encourage the agent to reach the goal quickly.

%-----------------------------------------------------------------------
\subsection{Proactive Global Planner for Waypoint Generation}
\label{section: global planner}
Our global planner generates waypoints for the local planner to follow. It assigns costs around each visible pedestrian according to a 2D Gaussian shape. Each pedestrian is inflated using a 2D Gaussian function, where closer pedestrians generate higher inflation based on their proximity to the robot. The pedestrian's heading is used to inflate more in the direction of the movement and less laterally by rotating the Gaussian in the heading direction, with standard deviation $\sigma_x > \sigma_y$. The center of the Gaussian is shifted by 2 units in the direction of the pedestrians's movement to create greater inlation in front of the pedestrian and less inflation behind them. This inflation serves as a forward projection of pedestrian movement, enabling the planner to proactively anticipate future positions and avoid potential collisions. According to the Gaussian shape, the costs of the 2D map is updated. A path planner then finds the shortest path to the goal location within this inflated map. This approach makes the path more suitable for avoiding future collisions. The Gaussian function is given below:
\begin{equation}
    g(p, \theta) = A \exp \left\{ -\frac{1}{2} \left [  \left(\frac{d_p \cos \theta}{\sigma _x} \right)^2 +  \left(\frac{d_p \sin \theta}{\sigma _y}\right)^2 \right ] \right\}
\end{equation}

\begin{figure}[h!]
\centering
\vspace*{0.5em}
\includegraphics[scale=0.35,angle=90]{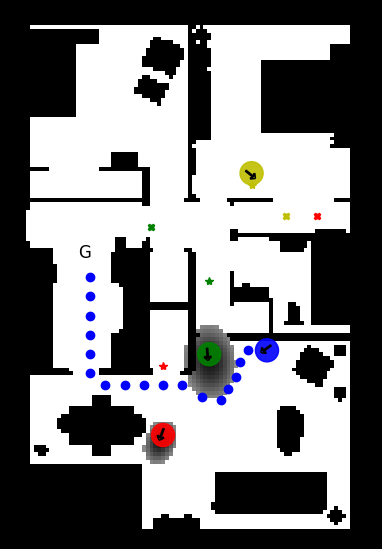}
\caption{
The robot (blue) detects two pedestrians (green and red). Our proactive global planner inflates the areas around the pedestrians on the map using a Gaussian (shaded regions). The closer pedestrian (green) is assigned a higher inflation compared to the farther one (red), reflecting their relative proximity. Additionally, more area is inflated in the direction of each pedestrian’s movement, ensuring that the planner accounts for their future positions. Using this inflated map, the planner recalculates a new, collision-free path that minimizes the risk of future interactions with the pedestrians.  
}
\vspace{-1em}
\label{fig:global_planner_fig}
\end{figure}

Here, $A = 1$ is the amplitude parameter which denotes the free space cost of 1 in the map. $p = (\Delta x, \Delta y)$ represents a position relative to 2 units ahead of the pedestrian's origin. $d_p$ is the distance between pedestrian origin and $p$. $\theta$ is the angle between pedestrian heading direction and $p$. $\sigma_x$ and $\sigma_y$ are the standard deviation:
\begin{align}
    \sigma_x &= r \times w_x,\ \ \sigma_y = r \times w_y \\
    r &= \frac{d_\text{ped}}{\text{max\_distance}}
\end{align}
Here, $w_x = 1m$ and $w_y = 0.7m$ are empirically chosen and r is the distance ratio which is proportional to the distance to the pedestrian from the agent, $d_\text{ped}$ and 
$g(p,\theta)$ is used to update the 2D costmap. Gaussian inflation of a pedestrian and planner path is shown in Fig. \ref{fig:global_planner_fig}.

%It inflates the positions of visible pedestrians based on their heading direction and distance from the agent. A path planner then finds the shortest path to the goal location within this inflated map. This approach makes the path more suitable for avoiding future collisions.

%The visible humans are inflated with a gaussian shape. The relative distance to a pedestrian is associated with the peak of the gaussian. The closer the humans, the more it's inflated. The heading is used to inflate more cells along the direction of the pedestrian. Less number of cells are inflated sideways. This inflation makes the planner aware of the human movement in advance. 

\section{Simulated Environment}
We created a multiagent 2D simulator for social navigation using indoor maps from 15 iGibson scenes \cite{shen2021igibson}. These scenes represent common indoor layouts, varying in size, rooms and objects. The scenes are divided into training, validation, and test sets, with 7, 2, and 6 scenes respectively.

The robot has a min/max linear and angular velocity of $\pm 0.5 \si{ \meter/\second}$ and $\pm \frac{\pi}{2} \si{\radian/\second}$ in the simulator. The action timestep is set to be $0.1$. The robot has a field of view of $90\degree$ and a depth sensor with a range of $5\si{\meter}$, which it can use to generate egocentric 2D partial map of the environment. Goal, waypoint and visible pedestrian positions, relative to the robot, are given in polar coordinates.  

Pedestrians are simulated in the environment using the ORCA simulator \cite{alonso2013optimal}. ORCA pedestrians are jointly optimized for their trajectories to avoid collision. By default, the robot is included in the ORCA model as an agent which makes the pedestrians cooperative to the robot assuming that the robot will also reciprocate in avoiding collision. This cooperation of the pedestrians mean that the pedestrians will not acively cause collision with the robot. However, since the robot has a different policy than ORCA, it can collide with the pedestrians. The pedestrians perception of the robot as another agent can be changed to simulate experiments where pedestrians are uncooperative toward the agent.

\section{Experiments and Results}
\subsection{Training}
We train the SAC policy for our local planner within the simulated environment. In each episode, the robot's start and destination positions are randomly initialized, with the shortest path distance ranging between 5 and 15 meters to capture a variety of navigational scenarios. The number of pedestrians per episode is randomly set between 1 and 5, and their start and goal positions are also randomized. When a pedestrian reaches its goal, a new random goal location is assigned. Training is conducted over 700 episodes per scene across 7 different scenes, with the scene changing after every 100 episodes. Every 200 episodes, we validate the model using 50 episodes per scene in two validation scenes, each containing 3 pedestrians. During training, all pedestrians were considered cooperative towards the robot, meaning they adjusted their behavior to avoid collisions.

\subsection{Evaluation}
For evaluation, we use 50 episodes per scene from the 6 test scenes. The evaluation metrics include \textbf{success rate (SR), average timesteps per episode (TS), average number of episodes with collisions (CO), and average number of episodes with timeouts (TO)}. Each planning strategy is tested in two sets of experiments: one with cooperative agents and another with non-cooperative agents, to assess the system's performance across different pedestrian behaviors. Evaluations were conducted using 3--10 pedestrians for each planning strategy. This range allowed us to test the performance of each in varying levels of crowd density.

Table \ref{tabl: comp} presents our evaluation results. The middle high-level column shows the average success rates (SR) for cooperative and uncooperative pedestrians across scenarios with 3 to 10 pedestrians. The rightmost column, labeled \textit{evaluation metrics}, provides the detailed evaluation metric scores, averaged for both cooperative and uncooperative pedestrians over the same range of pedestrian densities. We call this setting mixed scenarios. 

\subsection{Planning Strategies and Baselines}
We demonstrate the performance of our hyrbid system as well as that of each component---the reactive local planner and the proactive global planner---by comparing them against various baselines and evaluating them in different pedestrian behavior scenarios. Each are described below. 

\vspace{0.15em}\noindent \textbf{RRL+PPP} This is our system that combines the RRL planner for local navigation with the PPP for global waypoint generation. The PPP is also used during training to replan paths dynamically when needed. For the case of RRL$'$ + PPP, the PPP global planner was not used during training; instead, the waypoints remained fixed throughout the training process. This allows for a comparison of how the global planner's proactive behavior, introduced only during testing, influences the performance.

\vspace{0.15em}\noindent \textbf{RRL+A*} In this setup, we replace the global planner PPP with the A* algorithm while keeping RRL as the local planner. The A* planner is used during both training and evaluation. Similar to the previous case, RRL$'$+A* had waypoints fixed during training.

\vspace{0.15em}\noindent \textbf{RRL} In this variant, the waypoints are fixed at the start of each episode, and no dynamic replanning occurs during training or testing. 

\vspace{0.15em}\noindent \textbf{RL baseline} This baseline model follows a similar approach to a previous work~\cite{perez2021robot} with similar rewards. The waypoints are fixed at the beginning of the episode and the system does not incorporate orientation or pedestrian avoidance rewards, distinguishing it from our RRL model.

\vspace{0.15em}\noindent \textbf{DWA} The Dynamic Window Approach \cite{fox1997dynamic} is a popular non-learned local planner that takes into account the dynamics of the robot. It generates an admissible velocity space by evaluating possible linear and angular velocities and selects the best velocity based on a cost function that scores potential trajectories. The cost function is structured as follows:
\begin{align*}
    R_\text{DWA} = \alpha R_\text{heading} + \beta R_\text{speed} + \gamma R_\text{obstacle}
    % \text{cost}(v,\omega) &= \alpha \times \text{headingCost} + \beta \times \text{speedCost} \\ & + \gamma \times \text{obstacleCost}
\end{align*}
Here, $R_\text{heading}$ measures the angle difference between the trajectory endpoint and the local goal, $R_\text{speed}$ is the difference between the maximum possible speed and the current speed, $R_\text{obstacle}$ is related to the minimum distance between the trajectory and any obstacle.  $\alpha, \beta, \gamma$ are the weights and chosen to be 0.4, 1, and 0.1 respectively. 
%For the local planner, we compare against two baselines: the widely used non-learned local planner, Dynamic Window Approach (DWA), and an RL baseline similar to ours but without the orientation reward and pedestrian collision reward. For the global planner, we compare against the A* algorithm. 
\subsection{Evaluation of our Hybrid Planning System}
\label{sec: hybrid}
The results in Table \ref{tabl: comp} show that our hybrid system (RRL + PPP) outperforms the baselines in success rate (SR) across all crowd densities, especially in uncooperative pedestrian scenarios (65.70\% vs. 63.75\%). In mixed scenarios, our model had fewer collisions and personal space violations (PSV). However, for cooperative pedestrians, our global planner did not offer a significant advantage over RRL + A*, as the pedestrians naturally avoided collisions. PPP's cautious behavior resulted in slightly higher PSV (0.6\% vs. 0.9\%) and more timeouts (6.67\% vs. 4.92\%), as it caused the robot to take longer, more conservative paths.
%In this experiment, we compare our hybrid system (RRL + PPP) with all other baselines. The comparison is conducted with varying the number of pedestrians to evaluate performance under different crowd densities. As shown in Table \ref{tabl: comp}, our system consistently outperformed other methods in terms of success rate, even as the number of pedestrians increased. This demonstrates the robustness of our hybrid approach in complex environments with dynamic human interactions. When the pedestrian density reached higher levels, our hybrid system maintained a high success rate, highlighting its ability to adapt to crowded environments better than traditional or standalone DRL-based approaches. 
% \usepackage{tabularray}

\subsection{Evaluation of Local Planning Strategies}
\label{comp_local}
We evaluated the performance of the local planners (Table \ref{tabl: comp} Ablations: Local) by fixing the global planner (PPP) and deploying it with our reactive RL-based planner and two additional baselines: DWA and a reinforcement learning baseline. Note that both RRL$'$+PPP and RL Baseline are not trained with PPP directly during training. As presented in Table~\ref{tabl: comp}, RRL outperformed both DWA and the RL baseline across multiple metrics, including success rate, personal space violations, and collisions for all cases. Our choice of orientation and pedestrian avoidance rewards in RRL significantly enhances performance compared to the RL baseline, which lacks these rewards during training. The pedestrian avoidance reward increased awareness of nearby pedestrians, reducing collisions (17.30\% vs. 44.09\%) and personal space violations, thereby demonstrating better social compliance in the mixed scenario.

In contrast, DWA, while effective in simpler cases, experienced more collisions (23.43\%) because it treats pedestrians as static obstacles and only considers admissible velocities for the current timestep. This makes DWA less responsive, particularly in scenarios involving non-cooperative pedestrians. Although both the reactive planner and DWA used the same global planner with similar waypoints, DWA’s reliance on static assumptions led to a delayed response, resulting in more collisions, especially in dense pedestrian environments. This highlights the advantages of our dynamic, reward-driven approach in handling the complexity inherent in this domain.
% of real-world human behavior.

% % \usepackage[normalem]{ulem}
% % \usepackage{tabularray}
% \begin{table}[h]
% \centering
% \caption{Comparison of local planners when global planner is PPP}
% \label{tab:local_PPP}
% \begin{tblr}{
%   width = \linewidth,
%   colspec = {Q[167]Q[127]Q[117]Q[192]Q[173]Q[158]},
%   cell{1}{5} = {c},
%   hlines,
%   vlines,
% }
% \textbf{Planner}  & {\textbf{success rate}\\\textbf{~ ~ ~ ~ ~\%}} & {\textbf{avg steps}\\\textbf{per episode}} & {\textbf{personal space}\\\textbf{violatation steps,~\%}} & \textbf{episode w/~collision, \%} & \textbf{episode w/ timeout, \%} \\
% \uline{RRL}+PPP   & \textbf{76.09}                                & \textbf{195.6}                             & \textbf{1.96}                                             & \textbf{19.03}        &        4.3               \\
% RL Baseline + PPP & 54.41                                         & 190.66                                        & 3.78                                                        & 45.38                           & \textbf{0.2 }                         \\
% DWA + PPP         & 69.79                                         & 232.46                                     & 2.66                                                      & 24.79                        & 5.42                       
% \end{tblr}
% \end{table}

% \usepackage[normalem]{ulem}
% \usepackage{tabularray}
We conducted a second set of experiments (2nd section from Table \ref{tabl: comp}. Ablations: Local), with the waypoints fixed at the beginning of the episode. Fixing the waypoints makes the local planner solely responsible for collision avoidance. As shown in the results, RRL experienced higher success rate in all cases and significantly fewer collisions (25.11\% vs 44.66\%) compared to the RL baseline for the mixed scenario, further validating the effectiveness of our reward structure.

\subsection{Evaluation of Global Planning Strategies} 
We additionally compare global planners by fixing the local planner (RRL) and pairing it with each of two global planning strategies: PPP and A* (Table \ref{tabl: comp} Ablations: Global). Our global planner PPP achieves a higher success rate and fewer personal space violations in uncooperative scenarios compared to A*, which treats pedestrians as static obstacles. The proactive nature of PPP allows for more efficient, socially-aware navigation by continuously adjusting to pedestrian movements. By inflating more in the pedestrian’s direction of travel, PPP identifies safer paths, while A*'s static inflation approach often leads to suboptimal paths and a higher risk of collisions in dynamic environments. This underscores the importance of adaptive global planning in handling unpredictable human behavior. In the cooperative scenario, PPP did not show a significant advantage over A*, for reasons outlined in Sec.~\ref{sec: hybrid}, showing lower personal space violations (0.74\% vs. 1.13\%) but a higher timeout rate (7.13\% vs. 6.75\%).

\begin{table}
\vspace*{0.5em}
\centering
\caption{Comparison of different planners for social navigation}
\label{tabl: comp}
%\vspace*{0.5em}
\begin{tblr}{
  cells = {c},
  cell{1}{2} = {c=2}{},
  cell{1}{4} = {c=5}{},
  vline{2,4} = {1-16}{},
  hline{1-3,8,12,14} = {-}{},
  %columns = {colsep=2pt},
  column{1-3} = {colsep = 1.5pt},
  column{4-8}= {colsep = 3pt},
}
 & {Success \\~Rate} &   & {Evaluation Metrics (Avg of \\50\% CoOp \& 50\% UnCoOp ep)}  &  &    &    &  \\ \hline
{Planner}   & {CoOp} & {UnCoOp} & {SR} & {TS} & {PSV}  & {CO}   & {TO}   \\
RRL+PPP          & 92.01                 & \textbf{65.70}  & \textbf{78.86}                                           & 209.67 & \textbf{2.05} & \textbf{16.69} & 4.46 \\
RRL+A*            & \textbf{93.79}                 & 63.75  & 78.77                                           & 203.06 & 2.45 & 18.04 & 3.19 \\
RRL                & 86.05                 & 54.80  & 70.43                                           & 192.49 & 3.69 & 25.11 & 4.48 \\
RL Baseline        & 66.70                 & 43.29  & 55.00                                           & \textbf{190.26} & 3.75 & 44.66 & \textbf{0.33} \\
DWA                & 81.65                 & 49.13  & 70.39                                           & 229.62 & 2.83 & 23.43 & 6.17 \\ \hline
{Ablations: Local}   &   &    &    &   &     &    &   \\
RRL$^\prime$+PPP         & \textbf{90.92}                 & \textbf{64.30}  & \textbf{77.61}                                           & 209.51 & \textbf{2.11} & \textbf{17.30} & 5.11 \\
RL Baseline+PPP  & 66.96                 & 44.16  & 55.56                                           & \textbf{190.26} & 3.76 & 44.09 & \textbf{0.36} \\ 
DWA+PPP          & 81.65                 & 49.13  & 70.39                                           & 229.62 & 2.83 & 23.43 & 6.17 \\
RRL                & \textbf{86.05}                 & \textbf{54.80}  & \textbf{70.43}                                           & 192.49 & \textbf{3.69} & \textbf{25.11} & 4.48 \\
RL Baseline        & 66.70                 & 43.29  & 55.00                                           & \textbf{190.26} & 3.75 & 44.66 & \textbf{0.33} \\ \hline
{Ablations: Global} &   &   &   &   &   &   &      \\
RRL$^\prime$+PPP         & 90.92   & \textbf{64.30}  & \textbf{77.61}  & 209.51 & \textbf{2.11} & \textbf{17.30} & 5.11 \\
RRL$^\prime$+A*     & \textbf{91.33}     & 60.75  & 76.04  & \textbf{199.33} & 2.60 & 19.25 & \textbf{4.71} 

\end{tblr}
%\tablefootnote{CoOp: Cooperative pedestrians, UnCoOp: Uncooperative pedestrians}
\vspace{-1.5em}
\end{table}

\section{Conclusion}

In this work, we have introduced a hybrid framework for social navigation in complex indoor environments, combining the strengths of both classical planning and deep reinforcement learning (DRL) approaches. Our system, consisting of a proactive global planner and a reactive RL-based local planner, effectively handles the challenges posed by dynamic human interactions in cluttered spaces. By leveraging a global planner that dynamically updates waypoints based on pedestrian movements, and a local planner trained to avoid collisions while following these waypoints, we achieve a reliable solution that balances long-horizon planning with real-time responsiveness. Our experiments demonstrate that the proposed framework outperforms traditional methods and DRL-based baselines, particularly in challenging scenarios with humans agnostic to the presence of the robot. This work highlights the importance of modular hybrid approaches for enhancing the performance of social navigation in indoor environment. Future work could include incorporating additional pedestrian simulation models beyond ORCA to capture more diverse and realistic pedestrian behaviors and integrating group dynamics into the simulation to better reflect real-world pedestrian interactions. Another important direction is to introduce uncertainity in environments where the robot will not have prior knowledge of the layout.

\bibliographystyle{IEEEtran}
\bibliography{IEEEfull,main}

\end{document}